\documentclass[9pt]{article}

     \usepackage[final, nonatbib]{neurips_2019_ml4ps}

\usepackage[utf8]{inputenc} 
\usepackage[T1]{fontenc}    
\usepackage{hyperref}       
\usepackage{multirow}
\usepackage{url}            
\usepackage{amsmath}
\usepackage{graphicx}
\usepackage{booktabs}       
\usepackage{amsfonts}       
\usepackage{nicefrac}       
\usepackage{microtype}      
\usepackage{algorithmic,algorithm}
\usepackage[authoryear, round]{natbib}
\setcitestyle{authoryear,open={(},close={)}}
\hbadness=\maxdimen
\vbadness=\maxdimen
\title{Extracting more from boosted decision trees: A high energy physics case study}

\author{Vidhi Lalchand \\
  Department of Physics\\
  University of Cambridge, UK\\
  The Alan Turing Institute \\
  \texttt{vr308@cam.ac.uk}}

\begin{document}

\maketitle

\begin{abstract}
 
 Particle identification is one of the core tasks in the data analysis pipeline at the Large Hadron Collider (LHC). Statistically, this entails the identification of rare signal events buried in immense backgrounds which mimic the properties of the former. In machine learning parlance, particle identification represents a classification problem characterised by overlapping and imbalanced classes. Boosted decision trees (BDTs) have had tremendous success in the particle identification domain but more recently have been overshadowed by deep learning (DNNs) approaches. This work proposes an algorithm to extract more out of standard boosted decision trees by targeting their main weakness, susceptibility to overfitting. This novel construction harnesses the meta-learning techniques of boosting and bagging simultaneously and performs remarkably well on the ATLAS Higgs (H) to tau-tau data set \citep{open} which was the subject of the 2014 Higgs ML Challenge \citep{rm}. While the decay of Higgs to a pair of tau leptons was established in 2018 \citep{cms2017observation} at the 4.9$\sigma$ significance based on the 2016 data taking period, the 2014 public data set continues to serve as a benchmark data set to test the performance of supervised classification schemes. We show that the score achieved by the proposed algorithm is very close to the published winning score which leverages an ensemble of deep neural networks (DNNs). Although this paper focuses on a single application, it is expected that this simple and robust technique will find wider applications in high energy physics. 
 
\end{abstract}
\vspace{-15pt}
\section{Introduction}

Classifier learning is exacerbated in the presence of overlapping and imbalanced class distributions and specialized attention is needed to achieve reasonable performance. These pathologies are often witnessed in real world data sets such as those related to fraud detection, diagnosis of rare diseases or detecting previously unseen phenomena. In the binary classification context, overlapping classes are characterised by the presence of regions in the feature space with a high density of points belonging to both classes. Class imbalance on the other hand is characterised by significantly different prior class probabilities; there is a minority class with much fewer training samples than the majority class. A significant amount of literature targeting these challenges use ensemble classifiers to boost the performance of simple algorithms by using them iteratively in multiple stages or training multiple copies of the same algorithm on simulated re-sampled sets of the training data. The former is called \textit{boosting} and the latter is called \textit{bagging}. These meta-learning techniques are typically used independently as boosting targets reduction in bias and bagging targets reduction in variance. The algorithm proposed here is a variant of BDTs that jointly harness both boosting and bagging and perform remarkably well when the data exhibit class overlap and class imbalance. 
 
\section{Multivariate Techniques in HEP}

The search for interesting signatures in the extremely high volumes of data generated by LHC experiments requires more than just veteran physics ability \citep{radovic2018machine}. The use of machine learning techniques to classify particles and events has been mainstream for the past two decades. 

Neural nets had a timid start in high energy physics in the late 80s but since the advent of deep learning, NNs have started to be used more broadly \citep{baldi2014searching}. This is probably both due to an increased complexity of the data to be analysed and the demand for non-linear techniques. Deep learning is capable of discovering intricate structure in data by learning abstract representations of data that amplify the aspects needed for discrimination \citep{lecun2015deep}. While state-of-the-art results have been reported on several classification tasks, deep learning methods typically require a large amount of data to train and there is a high computational overload. BDTs have had tremendous success as a multivariate analysis tool in HEP. Pioneering works like \cite{roe2005boosted}, \cite{yang} have used carefully tuned BDTs for robust classification as a workhorse for particle identification.  

This short paper is organized as follows, section \ref{background} provides a brief summary on boosting and bagging, \ref{bxt} presents a variant of the vanilla boosted decision tree and discusses its performance edge in classification tasks which exhibit overlap and imbalance. Section \ref{case} summarises the Higgs to tau-tau channel dataset along with experimental results.

\section{Meta-Learning techniques: Boosting and Bagging}
\label{background}
One of the most fundamental expressions of boosting is the AdaBoost (Adaptive Boosting) algorithm \citep{scapire}; typically used with decision trees as the base learners. Boosting works by training base learners \footnote{can classify samples better than random guessing} to build a master learner which is significantly better at the task. The decision rule for the master learner is a weighted combination of the base learners outcomes and the weights are usually a function of the weak learners accuracy.
A boosted classifier with $J$ stages takes the form,
\begin{equation}
M_{J}(\mathbf{x}) = \sum_{j=1}^{J}\alpha_{j}h_{j}(\mathbf{x})
\end{equation}
where $h_{j}(\mathbf{x})$ are the base learners and the coefficients $\alpha_{j}$ denote the confidence in base learners. 
The training happens in stages and each training observation is associated with a weight. At the end of each stage the weight on misclassified samples is increased in a weight update step thereby allowing the classifier to focus on the hard to classify events in subsequent stages. At each iteration it improves accuracy by classifying few more points correctly without disrupting the correct classifications from previous stages. This provides some intuition about how boosting achieves low bias. However, this very property makes boosting prone to overfitting and makes it a high variance classifier. Boosting with several stages of tree learning can rapidly overfit as trees are low bias-high variance learners. Over-fitting control is one of the main challenges in using this technique. 

Bagging was a technique popularised by Leo Brieman who first published the idea in 1996 \citep{bagging}. The bagging technique helps stabilise classifiers by training a single classifier on multiple bootstrap samples (sampling with replacement) of the training data set. The size of the bootstrap sample usually is the same as the original data set. Due to sampling with replacement a bootstrap sample most likely has repeated samples, this increases the predictive force of the base learner on those samples. Brieman notes:

`\textit{The vital element is the instability of the prediction method. If perturbing the learning set can cause significant changes in the predictor constructed, then bagging can improve accuracy.}'

Random Forests exemplify the bagging principle. A random forest is a collection of tree learners each trained on a bootstrap sample of the training data. There are two sources of randomness in a random forest, (1) the formation of bootstrap samples and (2) a random selection of features considered at each node for splitting \citep{subspace}. An extremely randomized tree (ET) is one of the newer incarnations of the random forest method \citep{geurts2006extremely} which introduce further randomization in the tree construction process. While random forests attempt to create diversified trees by using a bootstrap samples, ETs work like random forests but take the randomization one step further by choosing a random split point at each node rather than searching for the best split. This ensures the creation of strongly \textbf{decorrelated} and \textbf{diverse} trees. From a computational view, ETs offer further advantage as they do not need to look for the most optimal split at each node. The algorithm proposed in section \ref{bxt} uses a bagged ensemble of ETs as the base learners within a boosting framework. This nested construction is able to counteract overfitting which is a fundamental weakness of classical BDTs. 

\section{Boosting Extremely Randomized Trees (BXT)}
\label{bxt}
At first the idea of using a randomized tree seems unintuitive as they are inferior to a tuned decision tree or a random forest which search for optimal split points, but it is precisely because of this that they serve as excellent weak learners. Bagging several extremely randomized trees alleviates overfitting and boosting ensures that their classification performance improves over stages. While it is true that BXT takes more stages of boosting to achieve the same performance of a classical BDT, it converges to a higher classification accuracy while the misclassification rate in a BDT saturates earlier. In the real world case study we show that this particular construction is superior to BDTs.
\begin{algorithm}[H]\small{
\caption{Boosting Extremely Randomized trees (BXT)}
\begin{algorithmic}[1]
\STATE \textbf{Training}:
\STATE \textbf{Input}: Training set $\mathbf{D}$, primitive tree learner $t$, initial weights $\{w_{i}^{1}\}_{i=1}^{N} = \frac{1}{N}$
\STATE \textbf{Output}: Master learner $M({\mathbf{x}}) = \sum_{j=1}^{J}\alpha_{j}T_{j}(\mathbf{x})$ where $T_{j}(\mathbf{x}) = agg{\{t_{1}(\mathbf{x}) \ldots t_{B}(\mathbf{x}) \}}_{j}$ is a collection of extremely randomized trees trained on $B$ bootstrap samples. $agg()$ refers to aggregation through majority vote. 
\item[]
\FORALL{$j$ = 1 \ldots $J$ (stages)}
\FORALL{$b$ = 1 \ldots $B$ (bootstraps)}
\STATE \textbf{Construct} a bootstrap sample ${D_{j}^{(b)}}$ (sample uniformly with replacement from $\mathbf{D}$)
\STATE \textbf{Fit} tree learner $t_{b}$ on sample $D_{j}^{(b)}$ using the Random Splitting Algorithm (see algorithm \ref{randomsplits})
\ENDFOR
\RETURN Set of extremely random fitted trees $\{t_{1} \ldots t_{B}\}_{j}$ together serving as the base learner $T_{j}$ for stage $j$. \\
\textbf{Compute} coefficient $\alpha_{j} = \frac{1}{2}\ln\bigg(\frac{N_{c}}{N-N_{c}}\bigg)$ where $N_{c}$ is the number of correctly classified samples for base learner $T_{j}$.\\
\textbf{Compute} weight update step \\
$w_{i}^{j+1} = w_{i}^{j}e^{\alpha_{j}\mathbf{1}(T_{j}(\mathbf{x}_{i}) \neq y_{i})}$\\
\ENDFOR
\item[]
\STATE \textbf{Testing}:
\STATE \textbf{Input}: Test set $\mathbf{L} = \{\mathbf{x}_{i}^{*}\}_{i=1}^{L}$, full set of parameters for the master learner $\{\{\alpha_{j}\}_{j=1}^J, \{T\}_{j=1}^{J}\}$ 
\STATE \textbf{Output}: Predicted labels $\{\hat{y_{i}}\}_{i=1}^{L}$ 
\item[]
\FORALL {$\mathbf{x}_{i} \in \mathbf{L}$}
\STATE \textbf{Predict} $\hat{y}_{i} = M(\mathbf{x_{i}}) = \sum_{j=1}^{J}\alpha_{j}T_{j}(\mathbf{x})$
\ENDFOR
\RETURN $\{\hat{y_{i}}\}_{i=1}^{L}$
\end{algorithmic}
\label{eforest}}
\end{algorithm}

\vspace{-20pt}
\section{Results on ATLAS Higgs to tau-tau data}
\label{case}

The ATLAS Higgs dataset from the CERN Open data portal \citep{open} has a total of 800K collision events (labelled $s$ or $b$) which have been simulated by the high energy ATLAS simulator. The simulator encompasses the current best understanding of the physics underlying the signal (events which generated a Higgs particle) and background decays. The signal events mimic properties of the background events as is known to happen in real events. The taxonomy of the dataset is given in appendix section \ref{tax}. For a more detailed description of the data set please refer to \cite{rm}. This data set is an example that encapsulates both class overlap and class imbalance. 


A typical binary classifier $h: \mathbb{R}^{d} \rightarrow \{b,s\}$ calculates a discriminant function $f(\mathbf{x}) \in \mathbb{R},\mathbf{x} \in \mathbb{R}^{d}$ which is a score giving small values to the negative class (background) and large values to the positive class (signal). One puts a threshold of choice $\theta$ (for this data set we choose a consistent cut-off threshold at the 85th percentile, a threshold chosen by physics experts) on the discriminant score and classifies all samples below the threshold as belonging to the negative class ($b$ or$-1$) and all samples with a score above the threshold as belonging to the positive class ($s$ or $+1$) or the selection region.

The goal of the classification exercise is to maximise a physics inspired objective called the \textit{Approximate Median Significance} (AMS) metric. One can think of the AMS as denoting a significance in terms of ($\sigma$). Given a binary classifier $h: \mathbb{R}^{d} \rightarrow \{b,s\}$, the AMS is given by, $\sqrt{2((\hat{s} + \hat{b})\ln(1 + \frac{\hat{s}}{\hat{b}})-\hat{s})}$ 
where $\hat{s}$ and $\hat{b}$ are the number of signal and background events in the selection region.

\begin{center}
\begin{table*}[ht]
\resizebox{\textwidth}{!}{
\begin{tabular}{l|c|c|c|c|c}
Learning Algorithm & Selection Region & False Positives & AMS ($\sigma$) & No. of learners & Training time (sec.)\\
\toprule
Boosted Decision Trees (BDT)  & 66789 & 5501& 3.73311$\sigma$ & 20 $\times$ 100 & 729.28 (CPU)\\
Boosted Extremely Random Trees (BXT)  & 66876 & 5413 & 3.79361$\sigma$ & 20 $\times$ 100 & 983.26 (CPU)\\
Ensemble of Neural Nets (DNN) (winning) & N/A & N/A & 3.8058$\sigma$ & 70 & 600 per NN (GPU) \\
\end{tabular}}
\caption{\small{Performance metrics of tree ensembles along with winning solution. Our score is only marginally lower than the winning score \protect\citep{melis} at the 85th percentile, and BXT is much computationally cheaper to train versus a DNN (several hours on a CPU).}}
\label{treeams}
\end{table*}
\end{center}





\vspace*{-40pt}
\begin{figure}[H]
    \includegraphics[scale=0.3]{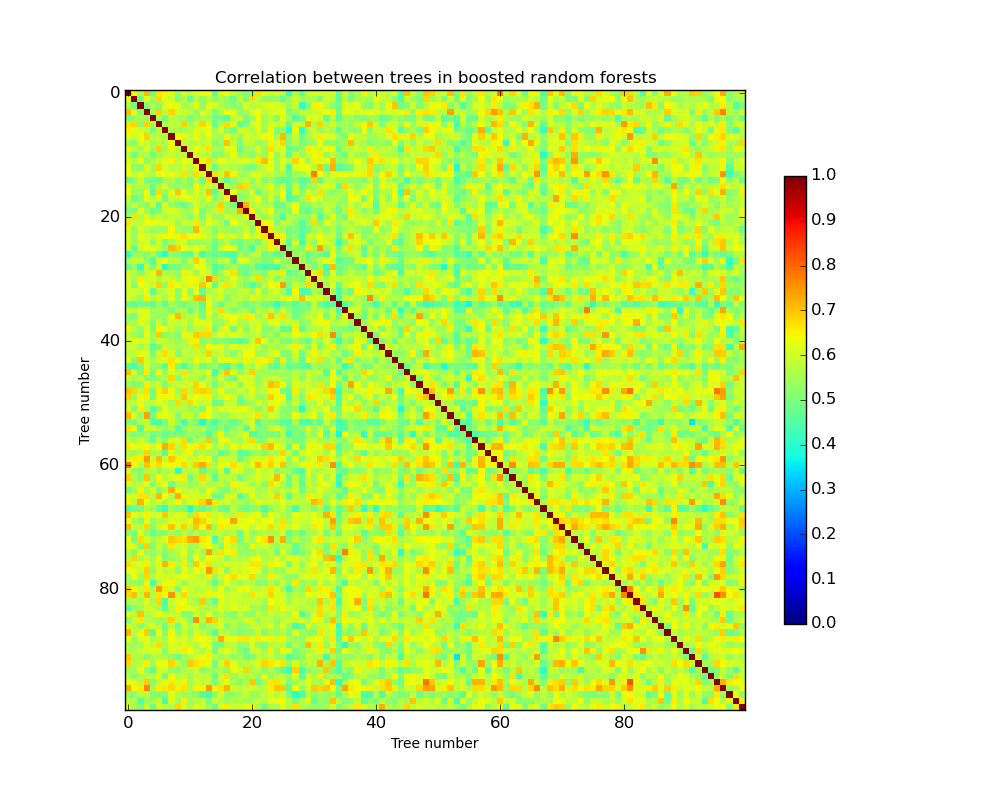}
    \includegraphics[scale=0.3]{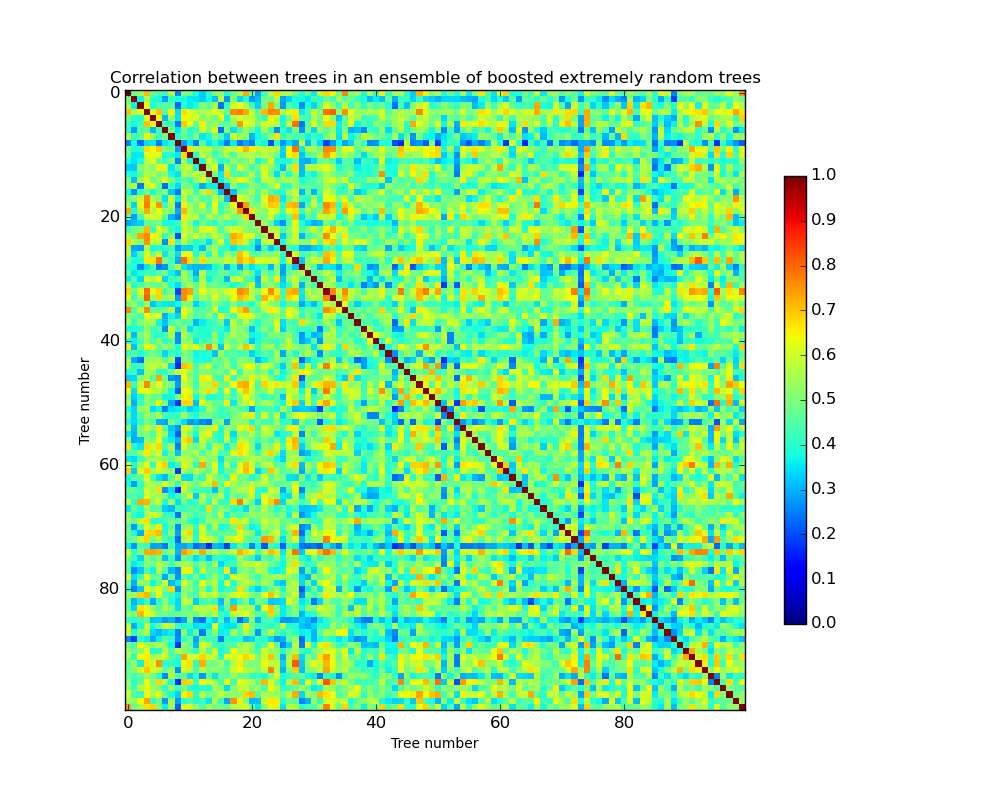}
    \vspace{-15pt}
    \caption{\small{Correlation among trees in a boosted ensemble of decision trees (BDTs, left) and extremely randomized trees (BXT, right). The figures depicts the correlation between 100 primitive trees picked at random from the 2000 trees that comprise the BDT and BXT model (20 stages of boosting $\times$ 100 trees in each tree ensemble). The higher level of de-correlation between the outputs of BXT versus the BDT indicates that diversity in ensembles enhances classification performance.}}
    \label{tc_brf}
\end{figure}
\vspace*{-20pt}

\begin{figure}[H]
    \includegraphics[scale=0.4]{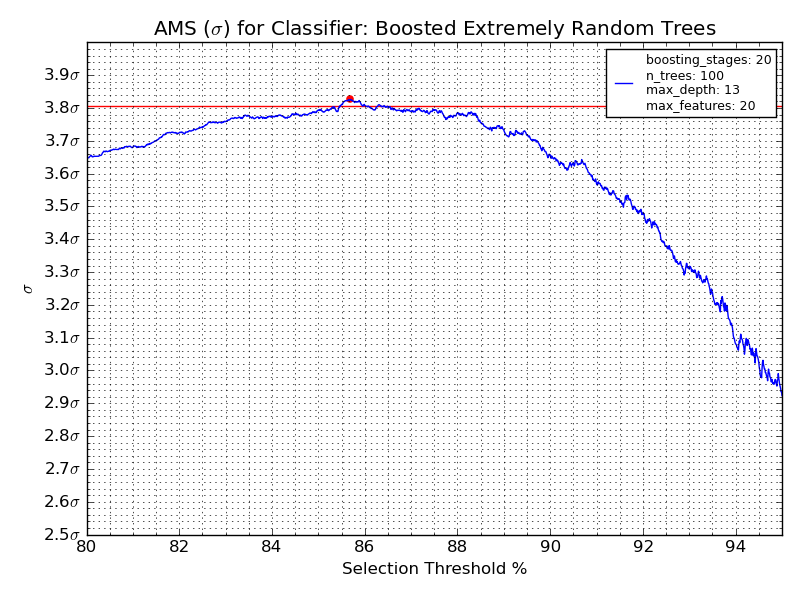}
    \includegraphics[width=0.6\textwidth]{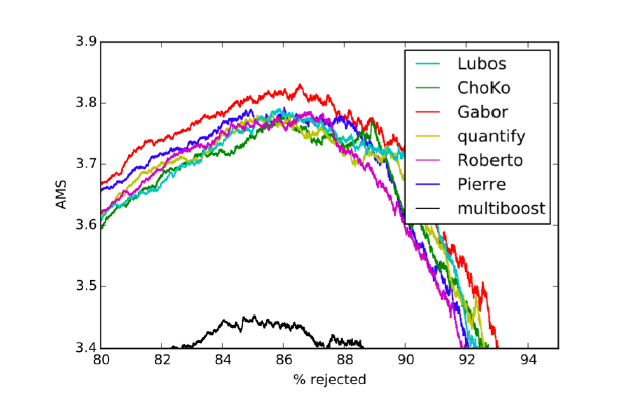}
        \vspace{-15pt}
    \caption{\small{AMS curves as a function of cut-off threshold for the selection region for our algorithm (left) and other dominant algorithms (winning one in red)}}
    \label{ams}
\end{figure}

\vspace*{-20pt}

\section{Discussion}
\label{close}

The compelling performance of a relatively simple algorithm based on boosting and extremely randomized trees hints  that we can extract a lot more mileage from primitive learning algorithms by embedding them in novel architectures. The BXT algorithm shines a light on the strength of diversity in ensemble learning; essentially by making predictors less deterministic and more diverse we can jointly improve classification performance. The general idea of using bagged learners within a boosting framework can be utilized with any component learner (not just trees). While deep learning approaches are powerful non-linear pattern extractors, their use might be overkill in many classification settings including high energy physics domains. Further, models based on primitive learners like trees are \textit{white-box} models; they are far more interpretable than DNNs.

In replacing BDTs with deep learning (DNNs), have we thrown the baby out with the bathwater? 

\subsubsection*{Acknowledgments}

The author would like to thank Anita Faul for guidance on this project and Christopher Lester for providing the dataset , physics motivation and useful discussions. VL is funded by The Alan Turing Institute Doctoral Studentship under the EPSRC grant EP/N510129/1.

\bibliography{main}

\section{Appendices}

\subsection{Random Splitting Algorithm}

The random split generating algorithm from \cite{geurts2006extremely} is summarized in \ref{randomsplits}. 

\begin{algorithm}[H]
\caption{Random Splitting algorithm}
\begin{algorithmic}[1]
\STATE \textbf{Input}: Training set $\mathbf{D}$
\STATE \textbf{Output}: A split $c_{j}$ which is a scalar value from a single feature vector of the training set $\mathbf{D}$  
\item[]
\STATE \textbf{Select} $K$ features ${a_{1}, \dots , a_{K}}$ at random from $\mathbf{D}$
\STATE \textbf{Select} $K$ splits $\{c_{1}, \dots , c_{K}\}$, one per feature for the $K$ features chosen in the previous step; each $c_{i}$ is selected at random from the range of the feature values $\forall i = 1, \dots K$  
\STATE Rank the splits $c_{i}$ by a criterion say $Q$ which gives a score $Q(\mathbf{D}, c_{i}) \in \mathbb{R}$ for each split. 
\STATE Select $c_{*} = max_{i=1 \dots K}Q(\mathbf{D}, c_{i})$  
\RETURN $c_{*}$
\end{algorithmic}
\label{randomsplits}
\end{algorithm}

\subsection{Taxonomy of the Higgs to tau-tau dataset}
\label{tax}
\begin{table}[H]
\begin{center}
\begin{tabular}{l|c|c|c}
Dataset & Total events & Background $|\mathcal{B}|$ &  Signal $|\mathcal{S}|$ \\
\toprule
Training & 250,000 & 164,333 & 85,667 \\
Cross Validation & 100,000 & 65,975 & 34,025\\
Testing & 450,000 & 296,317 & 153,683\\
\end{tabular}
\caption{ATLAS Breakdown of datasets for the learning task}
\label{break}
\end{center}
\end{table}

\subsection{Derivation of AdaBoost}

\label{deriv}

Consider a binary classification problem with input vectors $\mathbf{D} = \{(\mathbf{x}_{1},y_{1},w_{1}) \ldots (\mathbf{x}_{N},y_{N},w_{N})\}$ with binary class labels $y_{i} \in \{-1,+1\}, \forall i=1 \ldots N$. \\

At the start of the algorithm, the training data weights $\{w_{i}\}$ are initialized to $w_{i}^{(1)} = 1/N, \forall i=1 \ldots N$. A base classifier $h_{1}(\mathbf{x}): \mathbf{D} \rightarrow \{-1,+1\} $ that misclassifies the least number of training samples is chosen. Formally, $h_{1}(\mathbf{x})$ minimizes the weighted error function given by, 

\begin{equation}
R(h_{1}) = \sum_{i=1}^{N}w_{i}\mathbf{1}(h_{1}(\mathbf{x}) \neq y_{i})
\end{equation}

where $\mathbf{1}$ is the indicator function.  \\
 
After the first round of classification,  the coefficient $\alpha_{1}$ is computed that indicates the confidence in the classifier. It is chosen to minimize an exponential error metric given by,

\begin{align*}
E &= \sum_{i=1}^{N}e^{y_{i}\alpha_{1}h_{1}(\mathbf{x}_{i})}\\
&= \sum_{y_{i}\neq h_{1}(\mathbf{x}_{i})}e^{\alpha_{1}} + \sum_{y_{i} = h_{1}(\mathbf{x}_{i})}e^{-\alpha_{1}}
\end{align*}

\begin{align*}
\frac{dE}{d\alpha_{1}} &=  \sum_{y_{i}\neq h_{1}(\mathbf{x}_{i})}e^{\alpha_{1}} - \sum_{y_{i} = h_{1}(\mathbf{x}_{i})}e^{-\alpha_{1}} \\
&\Rightarrow \sum_{y_{i}\neq h_{1}(\mathbf{x}_{i})}e^{\alpha_{1}} = \sum_{y_{i} = h_{1}(\mathbf{x}_{i})}e^{-\alpha_{i}}\\
&\Rightarrow (N - N_{c})e^{\alpha_{1}} = N_{c}\frac{1}{e^{\alpha_{1}}}   \\
&\textrm{where $N_{c}$ is number of correctly classified samples by $h_{1}$}\\
&\Rightarrow e^{2\alpha_{1}} = \frac{N_{c}}{N - N_{c}}\\
&\Rightarrow \alpha_{1} = \frac{1}{2}\ln\bigg(\frac{N_{c}}{N-N_{c}}\bigg)
\end{align*}

Denoting $\epsilon_{1} = \dfrac{N - N_{c}}{N}$ as the error rate for $h_{1}$,

\begin{align}
\alpha_{1} = \frac{1}{2}\ln\bigg(\frac{1-\epsilon_{1}}{\epsilon_{1}}\bigg)
\label{alphas}
\end{align}

The weight update equation at each stage is given by, 

\begin{align*}
w_{i}^{(j+1)} = w_{i}^{(j)}e^{\alpha_{j}\mathbf{1}(h_{j}(\mathbf{x}_{i}) \neq y_{i})}
\end{align*}

The master learner $M_{J}(\mathbf{x})$ for a $J$ stage classifier is given by,

\begin{equation}
M_{J}(\mathbf{x}) =  \sum_{j=1}^{J}\alpha_{j}h_{j}(\mathbf{x})
\label{master}
\end{equation}

\end{document}